\documentclass[]{xiaomiev}

\usepackage[toc,page,header]{appendix}
\usepackage{minitoc}
\usepackage{solarized-light}
\usepackage{booktabs}
\usepackage{multirow}
\usepackage{graphicx}
\usepackage{array}
\usepackage{siunitx,array}
\usepackage[table]{xcolor}
\usepackage{makecell,array,booktabs}
\usepackage{makecell}
\usepackage{framed}
\usepackage{amsmath}
\usepackage{amssymb}
\usepackage{mathtools}
\usepackage{amsthm}
\usepackage{bbding}
\usepackage{hyperref}
\usepackage{amsmath} 
\usepackage{placeins}
\usepackage[colorinlistoftodos]{todonotes}
\usepackage{longtable}
\usepackage{hhline}
\usepackage{fancyvrb}
\usepackage{graphicx}
\usepackage[table]{xcolor}
\usepackage{booktabs}
\usepackage{float}
\usepackage{fvextra}
\usepackage{CJKutf8}
\usepackage{multicol}
\usepackage{float}
\usepackage{placeins}
\usepackage{cleveref}
\usepackage{tablefootnote}
\usepackage{threeparttable}
\usepackage{tabularx}
\usepackage{mdframed}
\usepackage{subcaption}
\usepackage[usestackEOL]{stackengine}
\usepackage[numbers]{natbib}
\newcommand{\commentout}[1]{}
\renewcommand{\paragraph}[1]{\noindent\textbf{#1.}\hspace*{1em}}
\newcommand{\na}{\cellcolor{gray!6}- -}
\usepackage{enumitem}
\usepackage{hyperref}
\setlist[itemize]{leftmargin=15pt}
\usepackage{siunitx,array}
\RequirePackage{xspace}
\makeatletter
\DeclareRobustCommand\onedot{\futurelet\@let@token\@onedot}
\def\@onedot{\ifx\@let@token.\else.\null\fi\xspace}

\makeatother

\title{OneVLA: A Unified Framework for Embodied Tasks}

\author{
  Lingfeng Zhang$^{1,2,3}$, 
  Xiaoshuai Hao$^{3,\dagger,\ddagger}$,
  Yingbo Tang$^{3,4}$,
  Lei Zhou$^{3}$,\\ 
  Shuyi Zhang$^{4}$,
  Jinkun Liu$^{1}$, 
  Hongsheng Li$^{1}$, 
  Chenhao Zhang$^{5}$,\\ 
  Qiang Zhang$^{6}$,
  Hangjun Ye$^{3}$,
  Xiaojun Liang$^{2}$,
  Long Chen$^{3}$,
  Wenbo Ding$^{1,\dagger}$
}

\affiliation[1]{Tsinghua University}
\affiliation[2]{Pengcheng Laboratory}
\affiliation[3]{Xiaomi EV}
\affiliation[4]{Institute of Automation, Chinese Academy of Sciences}
\affiliation[5]{Peking University}
\affiliation[6]{HKUST(GZ)}

\contribution[\dagger]{Corresponding authors}
\contribution[\ddagger]{Project leader}

\checkdata[GitHub]{\url{https://github.com/linglingxiansen/OneVLA}}
\correspondence{\email{haoxiaoshuai@xiaomi.com}, \email{ding.wenbo@sz.tsinghua.edu.cn}}

\abstract{
Navigation and manipulation are fundamental capabilities of embodied intelligence, enabling robots to interpret natural language commands and interact physically with their surroundings. 
However, current Vision-Language-Action (VLA) models remain constrained by task-specific architectures, specializing in either navigation or manipulation, which hinders the development of general-purpose robotic agents. 
To bridge this gap, we introduce \textbf{\textit{OneVLA}}, a unified architecture that integrates these distinct tasks into a single, cohesive framework.
Specifically, we design a unified action head capable of generating both navigation and manipulation actions without requiring task-specific variants.
Furthermore, we propose a multi-stage progressive training strategy—incorporating curated data construction and Chain-of-Thought (CoT) fine-tuning—that facilitates strong positive transfer and mutual reinforcement between the two domains. 
Extensive experiments in both simulated and real-world environments demonstrate that \textbf{\textit{OneVLA}} achieves state-of-the-art performance, significantly outperforming both specialized single-task and existing cross-task models. 
By unifying these core capabilities, \textbf{\textit{OneVLA}} paves the way for truly general-purpose robotic systems. 
The model and source code will be publicly released.}

\begin{document}
\maketitle
\vspace{-4pt}

\begin{figure}[!t]
    \centering
\includegraphics[width=\linewidth]{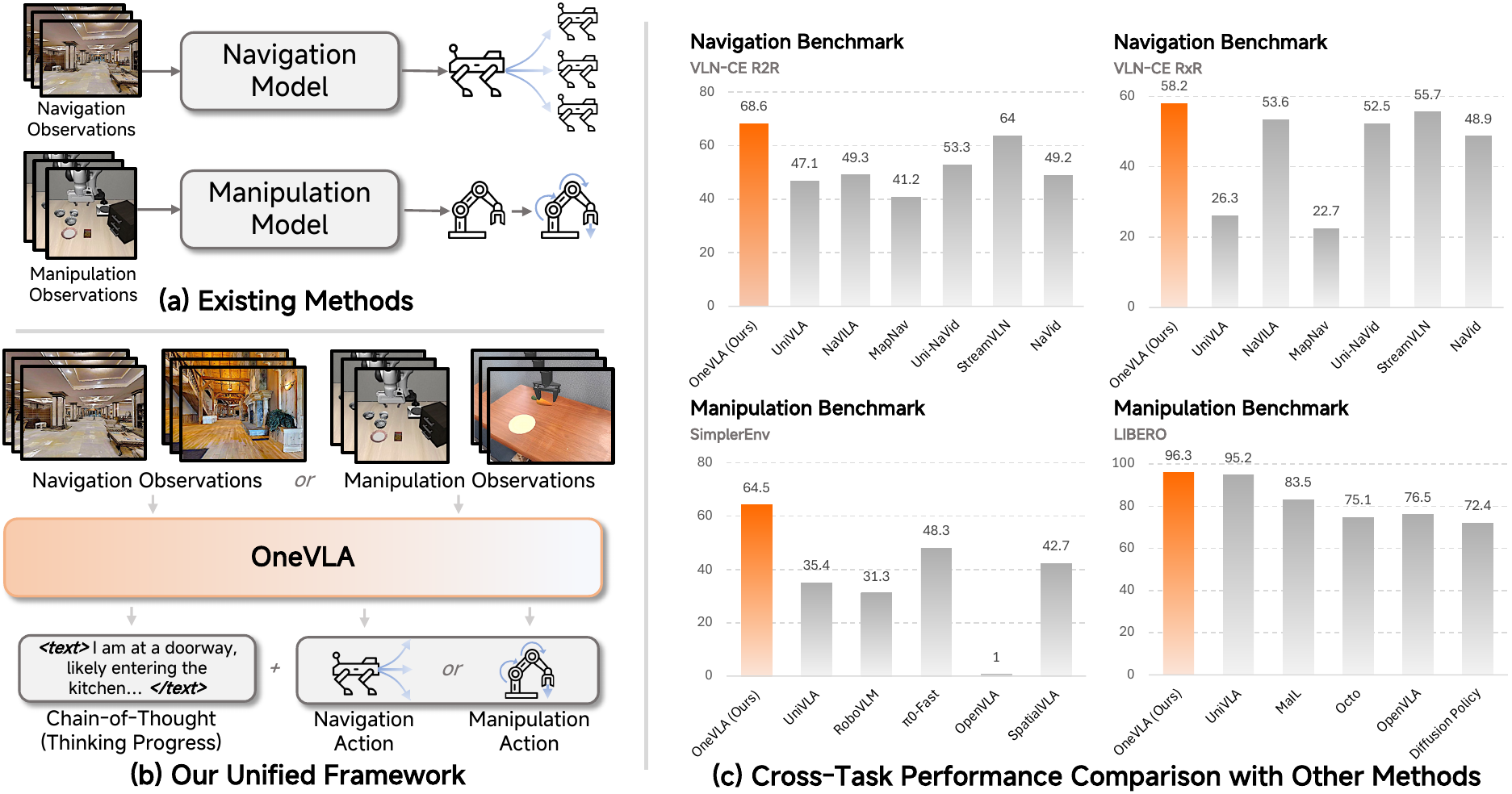}
    \caption{\textbf{Overview of \textbf{\textit{OneVLA}}.}
    (a) Previous methods require building separate models for navigation and manipulation tasks, limiting the transfer of cross-task knowledge. (b) Our \textbf{\textit{OneVLA}} proposes a \textbf{unified framework} that can handle navigation and manipulation tasks simultaneously in a single model, generating task-specific actions and general question-answering capabilities without task-specific model variants. (c) \textbf{\textit{OneVLA}} achieves state-of-the-art (SOTA) performance on multiple navigation and manipulation benchmarks, significantly outperforming single-task-specific models and existing cross-task methods.}
    \label{fig:model}
\end{figure}
\vspace{-1.5em}
\section{Introduction}

Embodied intelligence, which endows intelligent agents with the ability to physically interact with the real-world environment and humans, is a key pathway to achieving artificial general intelligence (AGI)~\cite{wuevaluating,liu2025embodied,tan2025roboos,mimoembodied,robosense2026,weatherbranch2026,sefmap2026}. 
In recent years, advancements in vision-language-action (VLA)~\cite{liu2025hybridvla,bu2025univla,wang2025unified,Hirose2025omnivla,onevl2026} models have significantly enhanced the performance of embodied intelligence tasks such as navigation and manipulation. 
VLA models combine the semantic understanding capabilities of vision-language models (VLMs)~\cite{li2024towards,goyal2025vla,bai2025qwen2} with perception and action generation capabilities. These models excel at translating natural language instructions into physical actions and have demonstrated impressive performance in various robotic tasks~\cite{black2025pi05,bai2025embodied,tang2025roboafford,roboaffordpp,spatialsky,toponav,socialnavmap2025,proactiverisk2025}.

However, current VLA models suffer from a fundamental architectural limitation that prevents them from becoming truly general-purpose robotic foundation models~\cite{uniact,bu2025univla}. 
Existing methods can be broadly categorized into two types: single-task VLA models focusing on either navigation or manipulation, and cross-task VLA models that attempt to handle multiple embodiments.
While cross-task models like UniVLA~\cite{bu2025univla} have made some progress in transferring knowledge across different robotic embodiments for navigation and manipulation tasks, they still require \textit{\textbf{task-specific model variants}}, and cannot seamlessly adapt to different embodied tasks.
This limitation raises two key questions: (1) \textit{Can we design a unified VLA architecture that can effectively perform navigation and manipulation tasks without task-specific model variants?} (2) \textit{Can these two fundamental embodied intelligence tasks mutually benefit each other during training, thereby improving overall performance?}

To address this challenge, we propose \textbf{\textit{OneVLA}}, a unified VLA architecture capable of handling both navigation and manipulation tasks within a single framework while overcoming the limitations of task-specific approaches.
As shown in Fig.~\ref{fig:model}, unlike previous methods that required \textbf{\textit{separate action heads}} or \textbf{\textit{specific model variants}} for different tasks, \textbf{\textit{OneVLA}} utilizes an innovative unified action output head as its core architectural innovation: concatenating the action dimensions of navigation and manipulation, thereby enabling the output of cross-task actions while maintaining independent loss calculations during training.
To equip the model with cross-task capabilities, we adopt a three-stage progressive training strategy. The first stage learns fundamental manipulation skills. The second stage incorporates navigation data to improve cross-task generalization and navigation performance. The third stage introduces Chain-of-Thought (CoT) data~\cite{videocot2025,reasoningmanifold2026,ivlr2026,onevl2026} for final refinement. In addition, we jointly train the backbone with a general vision–language objective, enabling shared vision–language representations and task-adaptive outputs.
Through extensive experiments in both simulated and real-world environments, we demonstrated that \textbf{\textit{OneVLA}} achieves state-of-the-art performance in both navigation and manipulation benchmarks, significantly outperforming single-task specialized models and cross-task models. More importantly, we provide empirical evidence that joint training on these two fundamental embodied tasks leads to mutual performance improvements, suggesting that a unified approach is not only more elegant architecturally but also superior in performance. 
Extensive experiments yielded two surprising findings: (1) \textit{A
unified VLA model can efficiently complete both
navigation and manipulation tasks without task-specific model variants,} and (2) \textit{Joint training of different embodied tasks can significantly improve the performance
of each other.}

Our main contributions are summarized as follows:
\begin{itemize}
    \item We propose \textbf{\textit{OneVLA}}, a unified VLA architecture that performs both navigation and manipulation tasks without task-specific model variants.
    \item We introduce an innovative unified action head and a multi-stage training strategy, enabling it to perform diverse embodied tasks and facilitating effective mutual enhancement between different task domains.
    \item Through comprehensive experiments, we demonstrate that \textbf{\textit{OneVLA}} achieves state-of-the-art performance on multiple benchmarks and provide evidence that navigation and manipulation tasks can mutually benefit from each other during mixed training.
    \item We break down previous architectural barriers between independent task domains, establishing a new paradigm for embodied intelligence and paving the way for general-purpose robotic systems capable of holistic environmental interaction.
\end{itemize}
\section{Related Work}

\textbf{VLA Models for Embodied Tasks.}
Vision-Language-Action (VLA) models~\citep{team2024octo, li2024cogact,hao2025tla,zhang2025vtla,liu2024robomamba,yang2025magma,song2025reconvla,oawam2026,ivlr2026,h2rbm2026} improve performance on embodied tasks by combining multimodal perception with motor control. However, recent methods generally focus on a single domain: in navigation, NaVid and Uni-NaVid~\citep{zhang2024navid,zhang2024uni} uses historical frame encoding to achieve temporally consistent navigation; MapNav~\citep{zhang2025mapnav,toponav,ascent2025,walkwithme2026,spatialsky,socialnavmap2025,proactiverisk2025} constructs explicit spatial maps to enhance path planning; and OmniVLA~\citep{Hirose2025omnivla} achieves cross-embodiment navigation by using goal-conditioned control. In manipulation, OpenVLA, RT-1-X, UniAct, etc.~\citep{kim2024openvla,o2023open,uniact} achieve strong generalization capabilities through large-scale pre-training on large scale dataset~\citep{o2023open,wu2024robomind,bu2025agibot}; $\pi_0$ model~\citep{black2024pi_0} employs a flow matching action head for precise control; and $\pi_{0.5}$~\citep{black2025pi05,oawam2026,ivlr2026,h2rbm2026,tang2025roboafford,roboaffordpp} further improves open-world performance through co-training on heterogeneous long-horizon data.
Despite these advances, all these VLA methods can only handle one task and lack cross-task capabilities. We propose \textbf{\textit{OneVLA}}, a unified VLA framework that can generate either navigation or manipulation actions within a single model, thus accomplishing both tasks.

\textbf{Cross-Embodiment Learning.}
Cross-embodiment learning aims to transfer knowledge to different robotic platforms~\cite{yang2024pushing,uniact,bu2025univla,mimoembodied,robosense2026,meshmimic2026}.
Previous work has primarily focused on cross-embodiment generalization for single tasks: RT-1-X~\cite{o2023open}, OpenVLA~\cite{kim2024openvla}, $\pi_{0.5}$~\cite{black2025pi05}, etc., use a shared motion space to implement operations by different robotic arms, while OmniVLA~\cite{Hirose2025omnivla} provides a general navigation strategy for heterogeneous robots. Octo~\cite{team2024octo}, UniAct~\cite{uniact}, etc., can adapt to different view and motion spaces, but are still limited to manipulation tasks. General strategies such as $\pi$ series~\cite{black2024pi_0,black2025pi05} can handle robotic manipulation or mobile manipulation tasks with different degrees of freedom, but mainly focus on manipulation tasks and lack fine-grained embodied navigation capabilities.
Recent cross-task studies, such as ECE~\cite{yang2024pushing}, CrossFormer~\cite{doshi2024scaling} and UniVLA~\cite{bu2025univla}, while possessing cross-task capabilities, still require \textbf{\textit{separate action heads}} or \textbf{\textit{specific model variants}} for different tasks.
These methods face significant limitations across embodiments and tasks. Our \textbf{\textit{OneVLA}} framework overcomes these challenges through a unified architecture and action head, enabling navigation or manipulation tasks to be performed effectively without different action heads or specific model variants.
\section{Methodology}
\begin{figure*}[!t]
\centering
\includegraphics[width=\textwidth]{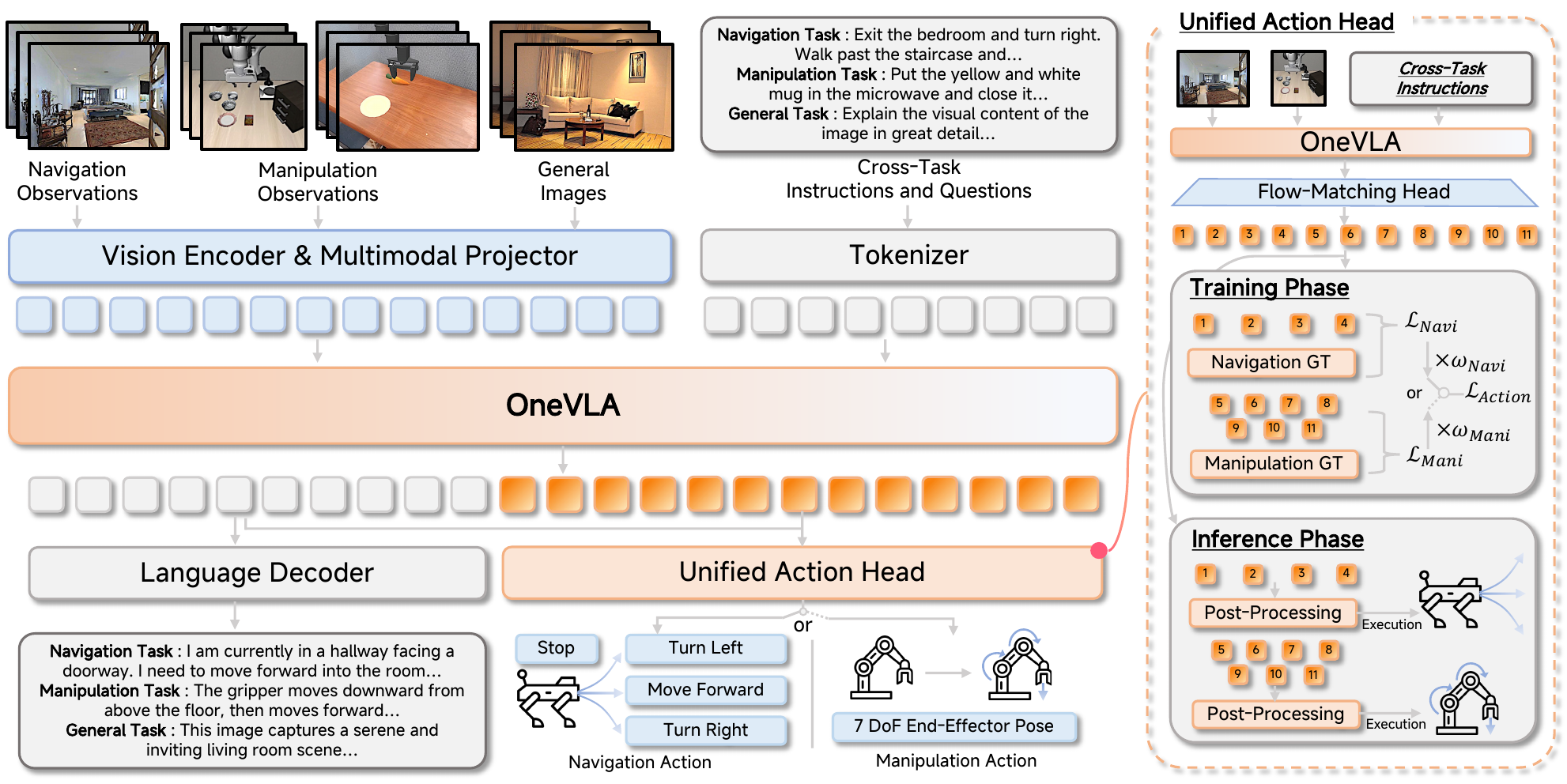}
\caption{
\textbf{Overall Framework of \textit{OneVLA}.}
For multimodal inputs in navigation or manipulation tasks (multi-view RGB images and natural language instructions), \textbf{\textit{OneVLA}} utilizes a unified vision encoder and tokenizer to extract cross-task vision-language features. A unified action head with a flow matching mechanism generates task-specific actions within an 11-dimensional unified action space: discrete navigation commands or continuous 7-DOF manipulation control. During training, we use task-specific weight masks to compute independent losses for different tasks, allowing the model to learn both tasks simultaneously without changing the architecture.
}
\vspace{-1em}
\label{framework}
\end{figure*}

\subsection{Overview}

We propose \textbf{\textit{OneVLA}}, a unified VLA framework capable of simultaneously generating both language responses and robotic actions for multiple embodied tasks. As shown in Fig.~\ref{framework}, at each time step $t$, given multimodal inputs including visual observations $o_t$ (multi-view RGB images), natural language instructions $l_t$, and optional robot state $r_t$, our model generates a dual output: a text response $y_t$ explaining the robot's understanding of the instruction; and a sequence of future actions $a_{t:t+T}$ for execution, in the following form:
$\text{OneVLA}: (o_t, l_t, r_t) \rightarrow (y_t, a_{t:t+T}).$
The action output $a_{t:t+T}$ can be adapted to different task domains without modifying the architecture. For embodied navigation tasks, the actions are processed into discrete movement commands represented as a probability distribution $a_{navi} = [p_{forward}, p_{turn\_left}, p_{turn\_right}, p_{stop}]$. For embodied manipulation tasks, the actions are continuous 7-DOF end-effector controls $a_{mani} = [\Delta x, \Delta y, \Delta z, \Delta \text{roll}, \Delta \text{pitch}, \Delta \text{yaw}, g]$. 
To achieve this dual output generation, we introduce special tokens \textless text\textgreater ...\textless /text\textgreater\textless action\textgreater ...\textless /action\textgreater, which explicitly distinguish the language generation phase and the action generation phase within a single forward pass. 

\subsection{OneVLA Architecture}
\textbf{Unified Vision-Language Encoder}
\textbf{\textit{OneVLA}} employs a unified vision-language backbone network to handle heterogeneous multimodal inputs. This encoder comprises two main components: a visual encoder for extracting spatiotemporal features from multi-view images; and a language model for fusing visual symbols and textual instructions into a unified representation space.

Given multi-view observations $o_t = \{I_1, I_2, ..., I_M\}$, where $M$ represents the number of camera views, each image $I_i \in \mathbb{R}^{H \times W \times 3}$ is first segmented into non-overlapping image patches of size $14 \times 14$. The visual encoder is a 32-layer Vision Transformer with a hidden dimension of $d_v = 1280$, which processes these image patches through self-attention layers with spatial merging (merge size = 2) to capture hierarchical visual features. The extracted visual features are then projected onto the hidden dimension $d_h = 2048$ of the language model through a linear projection layer:
\vspace{-0.3em}
\begin{equation}
\mathbf{V} = \text{Proj}(\text{VisionEncoder}(o_t)) \in \mathbb{R}^{N_v \times d_h},
\vspace{-0.3em}
\end{equation}
where $N_v$ represents the total number of visual symbols across all views.

Text instructions $l_t$ are tokenized and embedded into the same $d_h$ dimensional space. The language model is a 36-layer Transformer with 16 attention heads and 2 key-value heads (GQA), outputting a context-sensitive representation:
\vspace{-1em}
\begin{equation}
\mathbf{H} = \text{OneVLA}([\mathbf{V}; \mathbf{T}]) \in \mathbb{R}^{(N_v + N_t) \times d_h},
\vspace{-0.3em}
\end{equation}
where $N_t$ is the number of text tokens, and $\mathbf{H}$ serves as a unified multimodal representation for subsequent language and action generation. This architecture enables the model to perform cross-task visual-language feature extraction.

\textbf{Output Generation.}
To achieve simultaneous generation of language responses and robot actions, we introduce a token-based decoding mechanism that explicitly separates the two output modalities during a single forward propagation. This design allows the model to first reason about the task through language generation, and then generate the corresponding action sequence based on visual observation and the generated reasoning results.
We expand the vocabulary with four special tokens: \textless text\textgreater ...\textless /text\textgreater\textless action\textgreater ...\textless /action\textgreater. During inference, these tokens structure the autoregressive generation process:
\begin{equation}
\text{sequence} = \langle \text{text}\rangle \; y_t \; \langle/ \text{text}\rangle \; \langle \text{action}\rangle \; \hat{a}_{t:t+T} \; \langle/ \text{action}\rangle,
\end{equation}
where $y_t$ represents the text response notation explaining the robot's understanding, and $\hat{a}_{t:t+T}$ represents the discretized action notation. The complete hidden state $\mathbf{H} = \text{OneVLA}([\mathbf{V}; \mathbf{T}]) \in \mathbb{R}^{(N_v + N_t) \times d_h}$ provided by the vision-language encoder serves as the conditional vector for the action decoder, effectively capturing joint reasoning from both vision and language context.

\textbf{Unified Action Head.}
\label{actionhead}
The core innovation of \textbf{\textit{OneVLA}} lies in its unified action head, which seamlessly handles heterogeneous action spaces across different embodied tasks without requiring architectural modifications for specific tasks. Unlike previous works that required separate policy heads for navigation and manipulation, our method concatenates the action dimensions of these two tasks into a unified action space, while maintaining the independence of loss calculations during training:
\begin{equation}
    \mathbf{a}_{\text{unified}} = [\mathbf{a}_{\text{navi}} \mid \mathbf{a}_{\text{mani}}] \in \mathbb{R}^{11},
\end{equation}
where $\mathbf{a}_{\text{navi}} \in \mathbb{R}^{4}$ represents the navigation action, which is the probability distribution of discrete commands $[p_{\text{forward}}, p_{\text{turn left}}, p_{\text{turn right}}, p_{\text{stop}}]$; and $\mathbf{a}_{\text{mani}} \in \mathbb{R}^{7}$ represents the manipulation action, which is the continuous 7-DOF end-effector control $[\Delta x, \Delta y, \Delta z, \Delta \text{roll}, \Delta \text{pitch}, \Delta \text{yaw}, g]$.

Given the complete hidden state $\mathbf{H}$ provided by the visual language encoder, the action head employs a flow-matching diffusion process to predict future action sequences. Based on the flow-matching formula, we sample noise $\boldsymbol{\epsilon} \sim \mathcal{N}(\mathbf{0}, \mathbf{I})$ and interpolation time steps $t \sim \text{Beta}(\alpha, \beta)$, then construct a noisy trajectory $\mathbf{a}_t = (1 - t)\boldsymbol{\epsilon} + t\mathbf{a}_{\text{gt}}$, where $\mathbf{a}_{\text{gt}}$ is the true action sequence. The model learns to predict the velocity field $\mathbf{v}_\theta = \mathbf{a}_{\text{gt}} - \boldsymbol{\epsilon}$ by minimizing the flow-matching objective. The architecture comprises an action encoder that conditions the noisy action trajectory on time-steps using sinusoidal position encoding; a DiT backbone network with cross-attention layers that fuses the encoded action with the visual language features $\mathbf{H}$; and an action decoder that projects the diffused output into an 11-dimensional unified action space. During inference, the model uses Euler integrals to iteratively denoise N time steps: $\mathbf{a}_{t+\Delta t} = \mathbf{a}_t + \Delta t \cdot \mathbf{v}_\theta(\mathbf{a}_t, t, \mathbf{H})$, progressively refining the action prediction from random noise to the final output.

To address the semantic differences between discrete navigation actions and continuous manipulation control, we introduce dimension-specific loss weighting. For a training sample $i$ with task type $\tau_i \in \{\text{navi}, \text{mani}\}$, we define a task-specific weight mask to determine which dimensions contribute to the loss. The action loss $\mathcal{L}_{\text{action}}^i$ for each sample is calculated as follows:
\vspace{-0.5em}
\begin{equation}
    \mathcal{L}_{\text{action}}^i = \text{mean}(\mathbf{w}_{\tau_i} \odot (\hat{\mathbf{v}}_\theta^i - \mathbf{v}^i)^2),
    \vspace{-0.7em}
\end{equation}
 
where $\odot$ denotes element-wise multiplication, and $\mathbf{w}_{\tau_i} \in \{\mathbf{w}_{\text{navi}}, \mathbf{w}_{\text{mani}}\}$ is the task-specific weight mask. Key dimensions (such as the stopping action) are assigned higher weights to address class imbalance, while padding dimensions have near-zero weights to ensure they don't contribute to gradients during training. This design allows the model to learn two tasks simultaneously while respecting their unique action space features. During each forward propagation, model generates actions for a single task based on input instructions, with task-specific weight masks automatically selecting relevant action dimensions and applying appropriate supervision. 
This unified architecture eliminates the need for task-specific model variants or switching—task differentiation naturally arises from the learned cross-modal representations and weighted training objectives.

\textbf{Loss Function Design.}
The model is trained end-to-end using a composite loss function balancing the three objectives:
\begin{equation}
\mathcal{L}_{ \text{total}} = \lambda_{\text{vlm}} \mathcal{L}_{ \text{vlm}} + \lambda_{ \text{text}} \mathcal{L}_{ \text{text}} + \lambda_{ \text{action}} \mathcal{L}_{\text{action}},
\end{equation}
where $\lambda_{ \text{vlm}}$, $\lambda_{ \text{text}}$, and $\lambda_{ \text{action}}$ are weight coefficients used to control the contribution of each component.

To preserve the pre-trained vision language understanding capabilities and enhance cross-modal reasoning, we incorporated general vision language data (image descriptions, visual question answering) into the training of the model. For these samples, we only computed the VLM loss as:
\vspace{-0.5em}
\begin{equation}
    \mathcal{L}_{\text{vlm}} = -\sum_{i=1}^{N_{\text{vlm}}} \log p_\theta(y_{\text{vlm}}^i \mid \mathbf{o}_i, \mathbf{l}_i),
    \vspace{-0.5em}
\end{equation}
where $N_{\text{vlm}}$ is the number of general VQA samples in the batch, and $y_{\text{vlm}}^i$ is the target text response. This loss function applies standard cross-entropy to the generated tokens, enabling the backbone network to maintain strong vision-language alignment.

For embodied task samples (navigation and manipulation), we supervise the language reasoning phase to encourage explicit understanding of the task. We employ a standard next-token prediction objective that covers the entire reasoning sequence, including the boundary tokens (\textless text\textgreater, \textless/text\textgreater) as well as the internal content:
\begin{equation}
    \mathcal{L}_{\text{text}} = -\sum_{i=1}^{N_{\text{task}}} \sum_{j=1}^{|y_t^i|} \log p_\theta(y_{t,j}^i \mid \mathbf{H}^i, y_{t,<j}^i),
\end{equation}
where $N_{\text{task}}$ is the number of embodied task samples, and $|y_t^i|$ is the length of the inference text for sample $i$. $\mathcal{H}^i$ represents the multimodal hidden state of the encoder. This encourages the model to understand and express the action before generating it, thereby improving interpretability and reasoning capabilities.

For action generation, we compute the flow-matching loss using task-specific weight masks to accommodate the heterogeneous output spaces. The loss is defined as:
\vspace{-0.5em}
\begin{equation}
    \mathcal{L}_{\text{action}} = \frac{1}{N_{\text{task}}} \sum_{i=1}^{N_{\text{task}}} \text{mean}\left(\mathbf{w}_{\tau_i} \odot (\hat{\mathbf{v}}_\theta^i - \mathbf{v}^i)^2\right),
    \vspace{-0.5em}
\end{equation}
where $\mathbf{w}_{\tau_i} \in \{\mathbf{w}_{\text{navi}}, \mathbf{w}_{\text{mani}}\}$ is the task-specific weight mask, $\mathbf{v}^i = \mathbf{a}_{\text{gt}}^i - \boldsymbol{\epsilon}^i$ is the ground truth velocity field, and $\hat{\mathbf{v}}_\theta^i$ is the model's prediction. The weight mask ensures that only task-relevant action dimensions contribute gradients, enabling joint training on heterogeneous action spaces.

\begin{figure*}[!t]
\centering
\includegraphics[width=\textwidth]{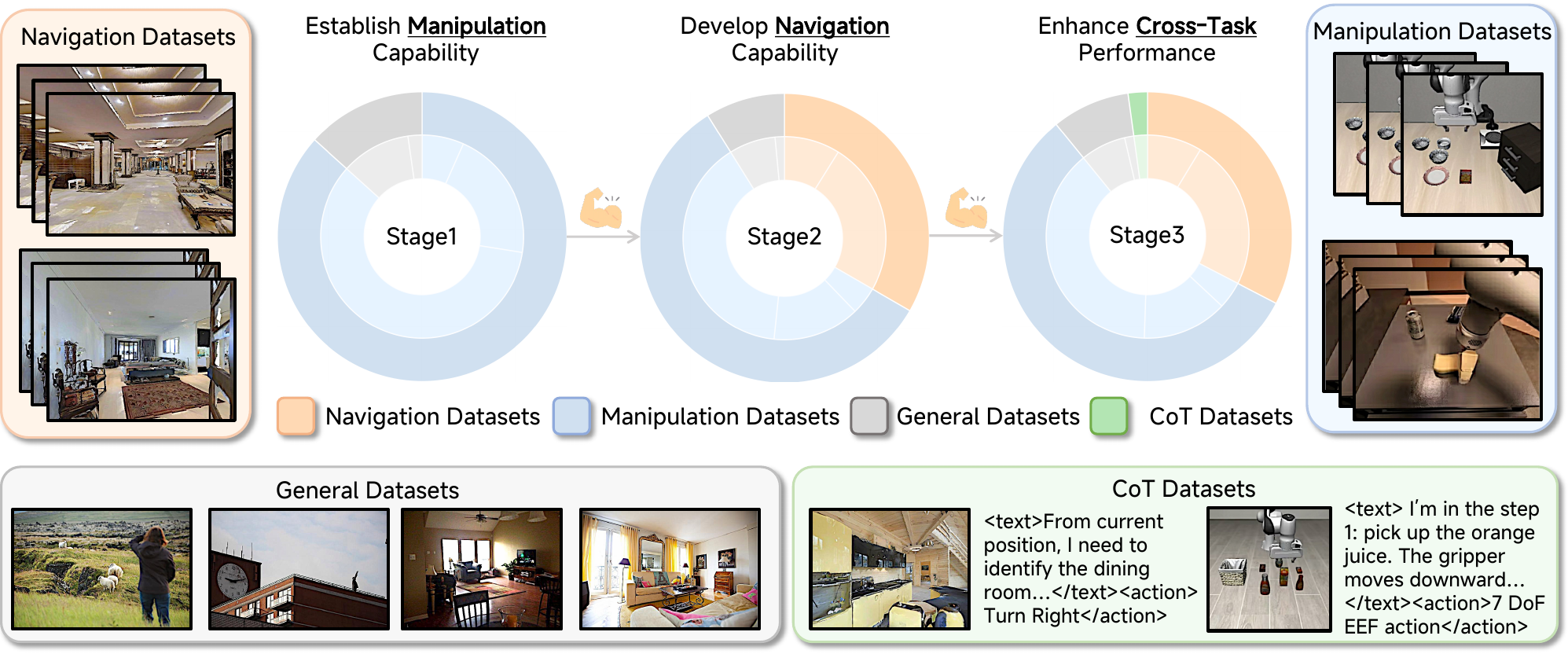}
\caption{
\textbf{Multi-Stage Training Strategy of \textit{OneVLA.}}
We propose a three-stage progressive training method: the first stage uses manipulation and general visual question answering (VQA) datasets to establish basic manipulation capabilities; the second stage introduces navigation data to develop navigation capabilities and achieve cross-task knowledge transfer through joint training; the third stage integrates Chain-of-Thought (CoT) datasets to enhance reasoning abilities and cross-task performance, ultimately yielding a unified model that performs well in both navigation and manipulation without requiring task-specific model variants.
}
\vspace{-1em}
\label{fig3}
\end{figure*}

\subsection{Multi-Stage Training Strategy}
As shown in Fig.~\ref{fig3}, to enable \textbf{\textit{OneVLA}} to effectively learn navigation and manipulation capabilities within a unified framework, we designed a three-stage progressive training strategy, gradually introducing task complexity and cross-task knowledge transfer.

\textbf{Establishing Manipulation Capabilities.} In the first stage, we focus on building robust manipulation skills, training solely using manipulation datasets and general vision-language datasets. The training data mixes manipulation task samples from an open-source cross-embodiment dataset with general visual question answer (VQA) samples. In this stage, the model learns language instructions through actions, translating them into continuous 7-DOF end-effector control while maintaining a strong understanding of visual language. The loss function for the first stage is:
\vspace{-0.4em}
\begin{equation}
    \mathcal{L}_{\text{stage1}} = \lambda_{\text{vlm}} \mathcal{L}_{\text{vlm}} + \lambda_{\text{action}} \mathcal{L}_{\text{action}}^{\text{mani}},
    \vspace{-0.4em}
\end{equation}
where the action loss $\mathcal{L}_{\text{action}}^{\text{mani}}$ uses a manipulation weight mask $\mathbf{w}_{\text{mani}}$ to supervise only the 7 degrees of freedom of the manipulation dimension.

\textbf{Developing Navigation Capabilities.} 
 In Stage 2, we introduce navigation data into the joint training while retaining manipulation and general VQA data. The key innovation of this phase is that the unified action head can now learn navigation and manipulation tasks simultaneously through task-specific weight masking, thereby enabling knowledge transfer between the two domains. The joint dataset consists of three parts: (1) navigation task samples from VLN dataset; (2) manipulation task samples from Stage 1; and (3) general VQA samples. The loss function becomes:
 \vspace{-0.3em}
\begin{equation}
    \mathcal{L}_{\text{stage2}} = \lambda_{\text{vlm}} \mathcal{L}_{\text{vlm}}  + \lambda_{\text{action}} (\mathcal{L}_{\text{action}}^{\text{navi}}+\mathcal{L}_{\text{action}}^{\text{mani}}),
    \vspace{-0.2em}
\end{equation}
 where $\mathcal{L}_{\text{action}}^{\text{navi}}$ and $\mathcal{L}_{\text{action}}^{\text{mani}}$ use $\mathbf{w}_{\text{navi}}$ and $\mathbf{w}_{\text{mani}}$ respectively. Critically, joint training allows the vision-language encoder to learn shared representations, thus benefiting both tasks.

\textbf{Enhancing Cross-Task Performance.} In the final stage, we introduce Chain-of-Thought (CoT) data to strengthen the model's reasoning capabilities and improve its decision-making quality across two tasks.
The training mixed dataset now contains all four data types: navigation, manipulation, general VQA, and CoT. The loss function is as follows:
\vspace{-0.3em}
\begin{equation}
    \mathcal{L}_{\text{stage3}} = \lambda_{\text{vlm}} \mathcal{L}_{\text{vlm}} + \lambda_{\text{text}} \mathcal{L}_{\text{text}} + \lambda_{\text{action}} (\mathcal{L}_{\text{action}}^{\text{navi}} + \mathcal{L}_{\text{action}}^{\text{mani}}),
    \vspace{-0.3em}
\end{equation}
where $\mathcal{L}_{\text{text}}$
supervises text generation on CoT samples. 

The multi-stage training strategy enables efficient development and knowledge transfer between embodied capabilities, ultimately generating a unified model that performs well in both navigation and manipulation \textit{\textbf{without requiring model variants for specific tasks}}.

\vspace{-1em}
\section{Experiments}

\begin{table*}[!t]
\centering
\caption{
\textbf{Performance Comparison of Models on Navigation and Manipulation Tasks.} The \textbf{\textit{Unified}} column indicates whether a model supports both navigation and manipulation tasks without task-specific model variants. UniVLA-Navi.$^*$ and UniVLA-Mani.$^*$ represent UniVLA's~\cite{bu2025univla} performance using \textbf{task-specific fine-tuned action heads} on the same training data for navigation and manipulation respectively. \textbf{``\colorbox{gray!10}{\na}''} indicates the model does not possess the capability to perform the specified task.}
\label{tab:zhubiao}
\renewcommand{\arraystretch}{1.5}
\resizebox{\textwidth}{!}{
\begin{tabular}{l|c|c|cc|ccccc}
\toprule
\multirow{2}{*}{\textbf{Method}} &\multirow{2}{*}{\textbf{Params}} &\multirow{2}{*}{\textbf{Unified}} & \multicolumn{2}{c|}{\textbf{Navigation (VLN)}} & \multicolumn{5}{c}{\textbf{Manipulation (SimplerEnv)}}  \\
\cmidrule(lr){4-5} \cmidrule(lr){6-10}
  &  & & R2R$\uparrow$ & RxR$\uparrow$ & \makecell{Put Sp.\\on Tow.$\uparrow$} & \makecell{Put Ca.\\on Pl.$\uparrow$} & \makecell{Stack Gr.\\on Ye.$\uparrow$} & \makecell{Put Egg.\\in Bask.$\uparrow$} & Avg.$\uparrow$ \\
\midrule
\multicolumn{10}{l}{\cellcolor{green!5}\textit{\textbf{Navigation VLA Models}}} \\
NaVid~\citep{zhang2024navid} &  7B & \XSolidBrush  &  49.2 & 48.9 & \na & \na & \na & \na & \na \\
Uni-NaVid~\citep{zhang2024uni} &  7B & \XSolidBrush  & 53.3 & 52.5 & \na & \na & \na & \na & \na \\
NaVILA~\citep{cheng2024navila} &  7B & \XSolidBrush  & 62.5 & 53.6 & \na & \na & \na & \na & \na \\
MapNav~\citep{zhang2025mapnav} & 7B & \XSolidBrush  & 41.2 & 22.7 & \na & \na & \na & \na & \na \\
StreamVLN-S.RGB~\citep{wei2025streamvln} & 7B & \XSolidBrush  & 64.0 & 55.7 & \na & \na & \na & \na & \na \\
\midrule
\multicolumn{10}{l}{\cellcolor{yellow!5}\textit{\textbf{Manipulation VLA Models}}} \\
RT-1-X~\citep{o2023open} & 35M  & \XSolidBrush  & \na & \na & 0.0 & 4.2 & 0.0 & 0.0 & 1.1 \\
Octo-Base~\citep{team2024octo} & 93M & \XSolidBrush  & \na & \na & 12.5 & 8.3 & 0.0 & 43.1 & 16.0 \\
$\pi_0$~\cite{black2024pi_0}& 3B  & \XSolidBrush  & \na & \na & 29.1 & 0.0 & 16.6 & 62.5 & 27.1 \\
$\pi_0$-Fast~\cite{pertsch2025fast}& 3B  & \XSolidBrush  & \na & \na & 29.1 & 21.9 & 10.8 & 66.6 & 48.3 \\

OpenVLA~\citep{kim2024openvla} & 7B & \XSolidBrush  & \na & \na & 0.0 & 0.0 & 0.0& 4.1 & 1.0 \\
OpenVLA-OFT~\citep{kim2025fine} & 7B & \XSolidBrush  & \na & \na & 34.2 & 30.0 & 30.0 & 72.5 & 41.8 \\
RoboVLM~\citep{li2024towards} & 2B  & \XSolidBrush  & \na & \na & 29.2 & 25.0 & 12.5 & 58.3 & 31.3 \\
SpatialVLA~\citep{qu2025spatialvla} & 7B  & \XSolidBrush  & \na & \na & 16.7 & 25.0 & 29.2 & 100.0 & 42.7 \\
Magma~\citep{yang2025magma} & 7B  & \XSolidBrush  & \na & \na &37.5 & 29.2 & 20.8 & 91.7 & 44.8 \\
\midrule
\multicolumn{10}{l}{\cellcolor{red!4}\textit{\textbf{Cross-Task VLA Models}}} \\
UniVLA-Navi.$^*$~\citep{bu2025univla} & 7B & \XSolidBrush &47.1 & 26.3  & \na & \na & \na & \na & \na \\
UniVLA-Mani.$^*$~\citep{bu2025univla} & 7B & \XSolidBrush & \na & \na & 37.5 & 33.3 & 4.2 & 66.7 & 35.4 \\
\rowcolor{orange!8}\textbf{OneVLA (Our Unified VLA Model)} & \textbf{3B} & \Checkmark & \textbf{68.6} & \textbf{58.2} & \textbf{87.5} & \textbf{37.5} & \textbf{50.0} & \textbf{83.3} & \textbf{64.5} \\
 \textit{vs. UniVLA~\cite{bu2025univla}} &  &  & \textcolor{blue}{\textbf{\textit{(+21.5)}}} & \textcolor{blue}{\textbf{\textit{(+31.9)}}} & \textcolor{blue}{\textbf{\textit{(+50.0)}}} & \textcolor{blue}{\textbf{\textit{(+4.2)}}} & \textcolor{blue}{\textbf{\textit{(+45.8)}}} & \textcolor{blue}{\textbf{\textit{(+16.6)}}} & \textcolor{blue}{\textbf{\textit{(+29.1)}}} \\
\bottomrule
\end{tabular}
}
\vspace{-1em}
\end{table*}

\subsection{Simulated Benchmarks}
To evaluate the performance of \textbf{\textit{OneVLA}} on navigation and manipulation tasks, we conducted extensive experiments in simulated environments using established benchmarks.

\textbf{Navigation Benchmarks.} For navigation evaluation, we employed the VLN-CE~\cite{krantz2020beyond} benchmark, which extends room-to-room (R2R)~\cite{krantz2020beyond} and room-to-room (RxR)~\cite{ku2020room} datasets to the continuous action space within the Habitat simulator~\cite{savva2019habitat}, requiring the agent to navigate to a target location following natural language instructions. We evaluated on the $\text{val}_\text{unseen}$ dataset, which contains previously unseen environments to test generalization capabilities. 
We used oracle success rate (OSR) as the primary metric, measuring the percentage of rounds in which the agent reached the target location within a specified distance threshold.

\textbf{Manipulation Benchmark.}
To evaluate manipulation performance, we used the SimplerEnv-Bridge~\cite{li24simpler} benchmark, which provides a realistic simulation environment capable of highly simulating the physical dynamics and visual appearance of the real world. We used a WidowX robotic arm equipped with the Bridge platform and focused on testing four representative tasks: (1) ``Put Spoon on Towel'', (2) ``Put Carrot on Plate'', (3) ``Stack Green Block on Yellow Block'', and (4) ``Place Eggplant in Yellow Basket''. 
Each task was run 24 times independently, and we report the success rate of each task as well as the average success rate of all four tasks. 

\vspace{-0.3em}

\subsection{Implementation Details}
\textbf{\textit{OneVLA}} uses Qwen2.5-VL-3B-Instruct~\cite{bai2025qwen2} as its unified vision-language backbone, with a hidden layer dimension of 2048. The unified action head adopts a DiT-B~\cite{peebles2023scalable} architecture with an 11-dimensional output space, which is composed of 4-dimensional navigation actions and 7-dimensional manipulation actions. We train the model using the AdamW~\cite{loshchilov2017decoupled} optimizer with learning rates of 4×10$^{-5}$ (base model), 1×10$^{-5}$ (vision-language interface), and 1×10$^{-4}$ (action model). 
Task sampling follows a 6:4 ratio of navigation data to manipulation data. The loss weights are set to $\lambda_{\text{action}}=1, \lambda_{\text{text}}=\lambda_{\text{vlm}}=0.1$.

\subsection{Comparison with SOTA Methods}
As shown in Tab.~\ref{tab:zhubiao}, \textbf{\textit{OneVLA}} achieves SOTA performance in both navigation and manipulation tasks benchmarks while maintaining a unified architecture without the need for task-specific model variants. Compared to UniVLA (7B)~\cite{bu2025univla}, the current most competitive cross-task baseline which requires task-specific action head fine-tuning, our model demonstrates significant performance improvements: \textbf{\textit{21.5}}\% increase on the VLN-CE R2R dataset (68.6\% vs. 47.1\%), \textbf{\textit{31.9}}\% increase on the RxR dataset (58.2\% vs. 26.3\%), and \textbf{\textit{29.1}}\% increase on the SimplerEnv dataset (64.5\% vs. 35.4\%). In addition to outperforming cross-task models, compared to single-task models, \textbf{\textit{OneVLA}} achieves \textbf{\textit{4.6}}\% and \textbf{\textit{2.5}}\% improvements over the latest SOTA navigation method StreamVLN (7B)~\cite{wei2025streamvln} on the R2R and RxR tasks, respectively; and \textbf{\textit{16.2}}\% improvement in average success rate on the SimplerEnv manipulation task compared to the latest SOTA method $\pi_0$-Fast~\cite{pertsch2025fast}. Notably, our \textbf{\textit{3B}} unified model outperforms all \textbf{\textit{7B}} task-specific models, demonstrating that effective architecture design and training strategies can achieve superior performance with significantly fewer parameters and without the need for task-specific model variants.

\subsection{Ablation Study}
\begin{table}[!t]
\centering
\caption{\textbf{Ablation Study on Multi-Stage Training.} UniVLA-Navi.$^*$ and UniVLA-Mani.$^*$ represent UniVLA's~\cite{bu2025univla} performance with \textbf{task-specific fine-tuned action heads} on the same training data for navigation and manipulation respectively.}
\resizebox{0.68\textwidth}{!}{
\begin{tabular}{l|c|c|c|c}

\toprule
\multirow{2}{*}{\textbf{Method}} & \multirow{2}{*}{\textbf{Params}} & \textbf{VLN} & \textbf{VLN} & \textbf{SimplerEnv} \\
& &\textbf{R2R}$\uparrow$ &\textbf{RxR}$\uparrow$&\textbf{Avg. SR}$\uparrow$ \\
\midrule
UniVLA-Navi.$^*$~\cite{bu2025univla} & 7B & 47.1 & 26.3 & \na \\
UniVLA-Mani.$^*$~\cite{bu2025univla} & 7B & \na & \na & 35.4 \\
\midrule
OneVLA (Single-Stage Training) & 3B & {56.1} & {42.9} & {46.2} \\
 \rowcolor{yellow!8}\textbf{OneVLA (Multi-Stage Training)} & 3B & \textbf{68.6} & \textbf{58.2} & \textbf{64.5} \\
\textit{vs. Single-Stage Training}& & \textcolor{blue}{\textbf{\textit{(+12.5)}}} & \textcolor{blue}{\textbf{\textit{(+15.3)}}} & \textcolor{blue}{\textbf{\textit{(+18.3)}}}\\

\bottomrule
\end{tabular}

}

\label{tab:ablation_stage}
\end{table}

\textbf{Effect of Multi-Stage Training.}
To validate the effectiveness of our proposed multi-stage training strategy, we conducted an ablation study comparing single-stage training with our three-stage progressive training method, where both approaches were trained on complete dataset for same steps. As shown in Tab.~\ref{tab:ablation_stage}, the multi-stage training strategy achieved significant improvements across all benchmarks: 12.5\% improvement on the VLN-CE R2R task (68.6\% vs. 56.1\%), a 15.3\% improvement on the RxR task (58.2\% vs. 42.9\%), and an 18.3\% improvement on the SimplerEnv manipulation task (64.5\% vs. 46.2\%).
These results demonstrate that gradually introducing task complexity leads to more effective knowledge transfer and learning compared to training with all data simultaneously from the beginning. Specifically, our strategy first establishes manipulation capabilities, then develops navigation skills, and finally leverages chain-of-thought data to enhance cross-task performance.

\textbf{Effect of Cross-Task Joint Training.}
We compare single-task training (single dataset) with joint training (both datasets in single stage) under the same training steps. As shown in Tab.~\ref{tab:ablation_cross_task}, joint training significantly improved performance compared to training each task separately: R2R performance improved by 7.3\% (56.1\% vs. 48.8\%), RxR performance improved by 9.3\% (42.9\% vs. 33.6\%), and SimplerEnv task performance improved by 5.5\% (46.2\% vs. 40.7\%). These results provide empirical evidence that navigation and manipulation tasks can mutually benefit from joint training, suggesting that shared vision-language representations learned from different embodied tasks contribute to better generalization and cross-task knowledge transfer.

\begin{figure}[!t]
    \centering
    \begin{minipage}{0.45\textwidth}
        \centering
        \vspace{1em}
        \includegraphics[width=\textwidth]{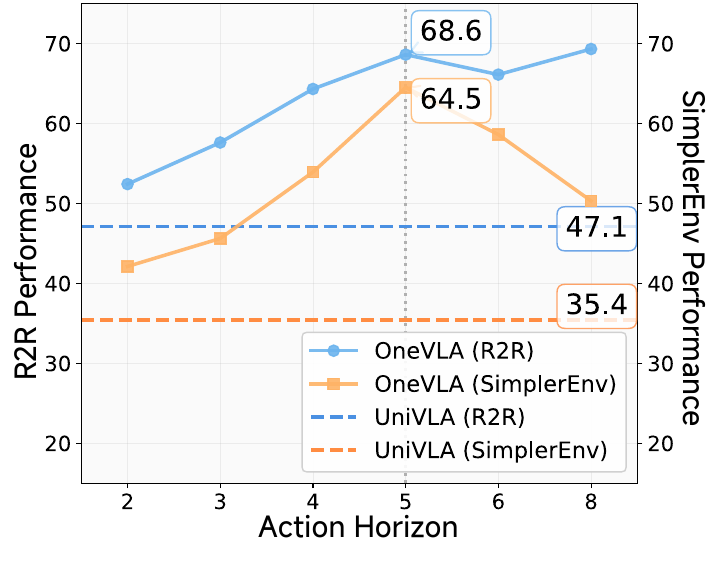}
        \captionsetup{aboveskip=-0.15em}
        \caption{Ablation Study on Different Action Horizon.}
        \label{fig:actionhorizon}
    \end{minipage}
    \hfill
    \begin{minipage}{0.45\textwidth}
        \centering
        \includegraphics[width=\textwidth]{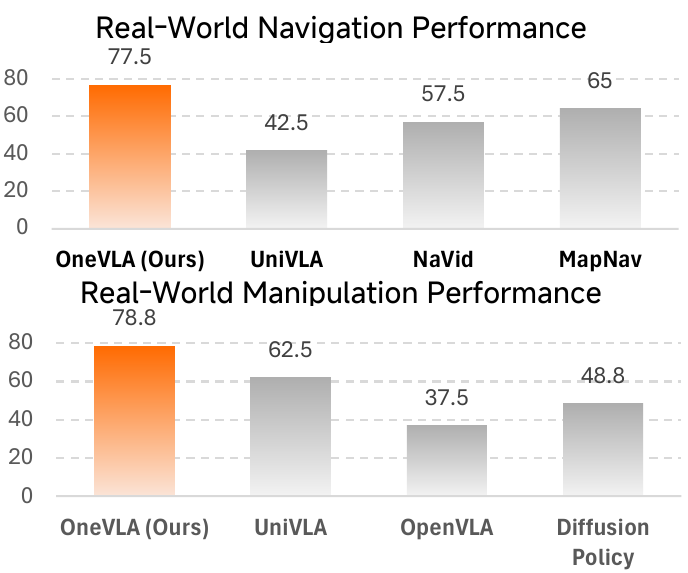}
        \captionsetup{aboveskip=-0.001em}
        \caption{Real-World Experimental Results.}
        \label{fig:realworld}
    \end{minipage}
\end{figure}

\begin{table}[!t]
\centering
\caption{\textbf{Ablation Study on Cross-Task Joint Training.} UniVLA-Navi.$^*$ and UniVLA-Mani.$^*$ represent UniVLA's~\cite{bu2025univla} performance with \textbf{task-specific fine-tuned action heads} on the same training data for navigation and manipulation respectively.}
\resizebox{0.68\textwidth}{!}{
\begin{tabular}{l|c|c|c|c}
\toprule
\multirow{2}{*}{\textbf{Method}} & \multirow{2}{*}{\textbf{Params}} & \textbf{VLN} & \textbf{VLN} & \textbf{SimplerEnv} \\
& & \textbf{R2R}$\uparrow$ & \textbf{RxR}$\uparrow$ & \textbf{Avg. SR}$\uparrow$\\
\midrule
UniVLA-Navi.$^*$~\cite{bu2025univla} & 7B & 47.1 & 26.3 & \na \\
UniVLA-Mani.$^*$~\cite{bu2025univla} & 7B & \na & \na & 35.4 \\
\midrule
OneVLA (Navi. Only) & 3B & 48.8 & 33.6 & \na \\
OneVLA (Mani. Only) & 3B & \na & \na & 40.7 \\

\midrule
\rowcolor{yellow!8}\textbf{OneVLA (Joint Training)} & 3B & \textbf{56.1} & \textbf{42.9} & \textbf{46.2} \\
\textit{vs. Single-Task Training} & & \textcolor{blue}{\textbf{\textit{(+7.3)}}} & \textcolor{blue}{\textbf{\textit{(+9.3)}}} & \textcolor{blue}{\textbf{\textit{(+5.5)}}}\\
\bottomrule
\end{tabular}
}

\label{tab:ablation_cross_task}
\end{table}
\textbf{Effect of Action Horizon.}
We conducted an ablation study to investigate the impact of action horizon on model performance in navigation and manipulation tasks. As shown in Fig.~\ref{fig:actionhorizon}, we evaluated action horizon ranging from 2 to 8 time steps while keeping all other hyperparameters constant. The results show that action horizon=5 achieves the best balance, yielding optimal performance on both R2R (68.6\%) and SimplerEnv (64.5\%) tasks. Shorter horizons (2-4 steps) lead to suboptimal performance due to insufficient temporal modeling. Conversely, excessively long horizons (6-8 steps) result in performance degradation, particularly in manipulation tasks, where SimplerEnv performance drops from 64.5\% (horizon=5) to 50.3\% (horizon=8). This is likely due to increased prediction difficulty and accumulated errors over longer time sequences.

\subsection{Real-World Experiments}
We conducted extensive real-world experiments for both navigation and manipulation tasks. For manipulation tasks, we designed four representative tasks and evaluated each task 20 times using a Franka Research 3 robotic arm, measuring the average success rate across all trials. For navigation tasks, we set up four different scenarios and tested each task 10 times using a mobile robot platform. As shown in Fig.~\ref{fig:realworld}, \textbf{\textit{OneVLA}} performed exceptionally well in real-world environments, significantly outperforming existing methods on both types of tasks. In manipulation tasks, our model achieved a success rate of 78.8\%, 16.3\% improvement over UniVLA~\cite{bu2025univla}. In navigation tasks, \textbf{\textit{OneVLA}} achieved a success rate of 77.5\%, 35.0\% improvement over UniVLA~\cite{bu2025univla}. These results demonstrate that \textbf{\textit{OneVLA}} not only performs well in simulated benchmarks but also effectively generalizes to real-world scenarios.

\subsection{Qualitative Analysis}

\begin{figure}[!t]
\centering
\includegraphics[width=\textwidth]{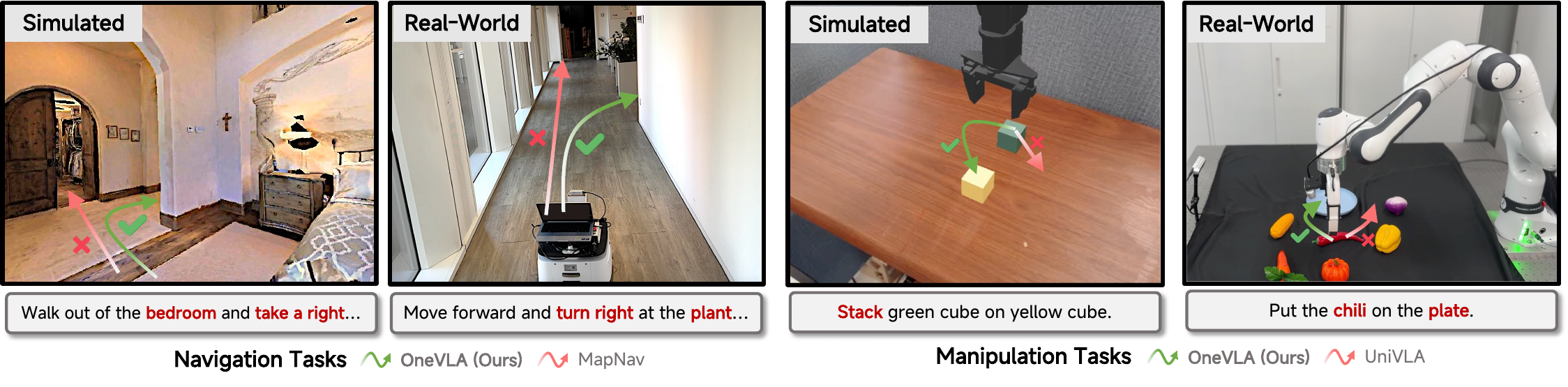}
\caption{
Qualitative Results in both Simulator and Real-World.
}
\vspace{-1.5em}
\label{qualitative}
\end{figure}

Fig.~\ref{qualitative} shows the qualitative results of \textbf{\textit{OneVLA}} in both simulated and real-world environments. In the simulated navigation task, our model successfully understands the instructions and generated corresponding movement commands. In the manipulation task, \textbf{\textit{OneVLA}} demonstrates precise object manipulation capabilities, accurately completing tasks such as stacking blocks and placing objects. More importantly, these capabilities effectively transferred to real-world scenarios. In real-world experiments, \textbf{\textit{OneVLA}} maintains robust performance in both navigation and manipulation tasks, successfully executing complex instructions using a real robotic platform. These qualitative results demonstrate that our unified framework can handle various embodied tasks in different environments.

\section{Conclusion}
In this paper, we introduce \textbf{\textit{OneVLA}}, the first unified vision-language-action framework designed to handle both navigation and manipulation tasks simultaneously with a single model, eliminating the need for task-specific variants. Using an innovative unified action head and a progressive multi-stage training strategy, \textbf{\textit{OneVLA}} achieves state-of-the-art performance across multiple benchmarks, showcasing the mutual benefits of joint training for navigation and manipulation. Extensive experiments in both simulated and real-world environments demonstrate that \textbf{\textit{OneVLA}} significantly outperforms existing single-task and cross-task methods, paving the way for general-purpose robotic systems capable of comprehensive environmental interaction.

\bibliographystyle{plainnat}
\bibliography{main}

\end{document}